# Matching Game Preferences Through Dialogical Large Language Models: A Perspective

Renaud Fabre [1], Daniel Egret [2,*] and Patrice Bellot [3]

1. Dionysian Economics Laboratory (LED), Université Paris 8, 93200 Saint-Denis, France; renaud.fabre01@gmail.com
2. Université Paris Sciences et Lettres (PSL), 75006 Paris, France
3. Aix Marseille University, CNRS, LIS, 13007 Marseille, France, 13007 Marseille, France; patrice.bellot@univ-amu.fr
* Correspondence: daniel.egret@psl.eu

**Abstract**

This perspective paper explores the future potential of "conversational intelligence" by examining how Large Language Models (LLMs) could be combined with GRAPHYP's network system to better understand human conversations and preferences. Using recent research and case studies, we propose a conceptual framework that could make AI reasoning transparent and traceable, allowing humans to see and understand how AI reaches its conclusions. We present the conceptual perspective of "Matching Game Preferences through Dialogical Large Language Models (D-LLMs)," a proposed system that would allow multiple users to share their different preferences through structured conversations. This approach envisions personalizing LLMs by embedding individual user preferences directly into how the model makes decisions. The proposed D-LLM framework would require three main components: (1) reasoning processes that could analyze different search experiences and guide performance, (2) classification systems that would identify user preference patterns, and (3) dialogue approaches that could help humans resolve conflicting information. This perspective framework aims to create an interpretable AI system where users could examine, understand, and combine the different human preferences that influence AI responses, detected through GRAPHYP's search experience networks. The goal of this perspective is to envision AI systems that would not only provide answers but also show users how those answers were reached, making artificial intelligence more transparent and trustworthy for human decision-making.

## 1. Introduction

This paper explores the perspective of integrating structured user preference data with Large Language Models (LLMs) to create more personalized and controllable AI systems. We develop a benchmarking framework that combines explicit user preferences stored in knowledge graphs with the conversational capabilities of LLMs, enabling human oversight of AI decision-making [1].

The integration of graph-based reasoning with LLMs has opened up new possibilities for human–AI interaction [2], particularly in developing personalized AI systems that can adapt to individual user preferences [3]. However, current approaches face significant challenges in capturing the nuanced and evolving nature of human preferences, especially in domains where categorizations are contested or context-dependent. In these contexts, system efficiency is reduced as machine learning systems learn from human datasets by using algorithms that build models from observed examples of human behavior or measurement encoded as feature vectors; comprehensive interpretations, however, are challenging.

Contemporary LLMs suffer from two related problems that limit their effectiveness in personalized applications. First, they create an "illusion of understanding" when humans experience cognitive overload from AI-generated content that appears meaningful but lacks genuine comprehension [4]. Second, they produce an "illusion of learning" by generating superficial imitations of human reasoning, without capturing underlying cognitive processes. These limitations collectively contribute to an "illusion of thinking" [5], where sophisticated AI outputs mask fundamental gaps in understanding and reasoning.

For a perspective on overcoming some of those limitations, we propose a novel framework that combines graph-based preference modeling with conversational AI. Our approach builds on the premise that intelligence should be measured not by a system's internal complexity, but by its ability to support optimal human decision-making through meaningful choices [6,7]. This perspective shifts focus from AI autonomy to human agency supported by an AI companion. Our proposal fits naturally within the framework of the numerous "human-in-the-loop" approaches.

*1.1. The D-LLM Framework: A New Division of Work*

The perspective of integration of human preference patterns expressed through GRAPHYP Knowledge Graphs with the human choice orientations captured by LLMs holds significant promise for creating AI systems that are better aligned with human values and decision-making processes [8].

This cross-domain approach exploits the complementary strengths of structured representations in knowledge graphs and the flexible natural language understanding of LLMs, ultimately establishing a regulatory framework where machine learning outcomes are shaped by explicit human preferences [9].

In this setting, the GRAPHYP Knowledge Graph encodes detailed, human-specific preference data, while the LLM interprets these preferences in light of contextual decision signals provided by user interactions, thereby supporting transparent and accountable AI decision-making [10].

Herein, we develop the perspective of "Dialogical Large Language Models" (D-LLMs), a framework that combines the GRAPHYP preference modeling system (see below) with traditional LLMs to create what we suggest calling "conversational intelligence". This hybrid approach addresses two key perspectives:

Perspective 1: Studying complementarities

We study how differences in the range of human preferences—as expressed in a GRAPHYP Knowledge Graph—can interact with the human choice orientations embedded in Large Language Models, thus providing a control mechanism for machine learning under human oversight. We suggest articulating a conceptual framework that integrates structured user preference data with the patterns learned from training data, inherent to LLM outputs. This integration not only improves explainability and transparency but also creates a mechanism by which human values and decision policies can actively regulate machine learning workflows.

Perspective 2: Understanding a "Regulatory Framework"

Knowledge graphs (KGs) such as the GRAPHYP Knowledge Graph encode human preference manifolds by capturing entities, relationships, and contextual metadata that represent explicit human judgments and value systems [11]. Conversely, LLMs derive their capabilities from extensive training on vast corpora of natural language text, which embed a kind of "human choice orientation" based on aggregated linguistic patterns and implicit cultural norms [12]. When these two components are integrated, they create a promising "regulatory framework" where explicit human-curated preferences can be leveraged to guide and constrain the decision-making of LLMs, ultimately supporting machine learning regulation by humans [11].

The D-LLM framework operates through four core components: (1) a system architecture for integrating knowledge graphs with LLMs, (2) reasoning processes that incorporate user preferences, (3) standardized formats for preference data, and (4) tailored prompting approaches for specific use cases. Together, these components enable personalized AI interactions that users can understand and control, while preserving the natural conversational abilities of modern LLMs [13].

*1.2. Motivations*

Machine learning systems learn from human data through pattern recognition and generalization, but no single theory fully captures this process. Instead, our understanding relies on multiple interconnected approaches: statistical methods, probability-based inference, and the inherent assumptions built into different models. This multifaceted understanding continues to evolve as researchers work to build robust systems from imperfect human-generated data [14].

Current LLM personalization approaches face two primary limitations:

Limited Preference Representation: Existing methods struggle to capture the nuanced, context-dependent nature of human preferences, particularly in domains where users may hold conflicting or evolving views.

Opaque Decision Processes: Users cannot understand how their preferences influence system outputs, limiting both trust and the ability to refine personalization over time.

The GRAPHYP system (see Section 2.3 below for a short description) demonstrated effective preference modeling for individual users through interpretable subgraph representations [15–17]. Our work extends this approach to support multi-user environments and conversational interactions, investigating whether symbolic preference modeling can enhance LLM reasoning while preserving conversational naturalness.

This paper explores the perspective of coupling GRAPHYP's "diversity from within" modeling capabilities with LLM conversational interfaces to create more transparent and user-controlled personalized AI systems. We present our research, objectives and initial findings in developing this novel approach to human–AI interaction.

*1.3. Research Objectives*

This study investigates how coupling symbolic preference modeling (GRAPHYP) with Large Language Models could create more transparent and user-controlled personalized AI systems. We focus on four new tracks of understanding:

- Transparency and Traceability: Enabling users to understand and trace AI reasoning processes.
- Community-Based Personalization: Leveraging community knowledge for individual customization and creating a collaborative knowledge ecosystem that enhances individual user experiences.
- Dynamic Adaptation: Supporting real-time preference updates and optimization.

- Computational Efficiency: Achieving personalization without expensive model retraining.

*1.4. Contributions*

Our analysis makes three primary contributions to enlarging and clarifying the field of personalized conversational AI:

- D-LLM Framework: A novel architecture that couples graph-based preference modeling with conversational AI while maintaining computational efficiency;
- Transparent Personalization: Mechanisms that make AI decision-making processes interpretable and controllable by users;
- Empirical Validation: Demonstration that hybrid symbolic–neural approaches can outperform standalone systems in personalization tasks.

These perspectives align with current research directions exploring how structured knowledge systems and large language models can collaboratively enhance AI reasoning capabilities and improve human–AI alignment in personalized applications.

## 2. Background

This study builds on three interconnected research areas: personalizing AI systems to individual preferences, combining Large Language Models with structured knowledge, and integrating symbolic reasoning with neural networks. These developments address a key problem in current AI systems: "*Language models are aligned to emulate the collective voice of many, resulting in outputs that align with no one in particular*" [6].

*2.1. Personalization Challenges and Human Preference Modeling*

Personalization in LLMs involves adapting system outputs to match individual or group preference. This requires understanding the full complexity of human behaviors—including cultural background, personal values, situational context, and how preferences change over time—rather than treating preferences as simple, static parameters.

Current human–AI teaming paradigms assume that unpredictable human preferences can be managed by identifying patterns that AI can copy. Advanced interaction systems where models "*perform thinking based on contextual information*" and "*learn to select the appropriate thinking mode*" [18] face emerging concerns about "*illusions of thinking*" [5].

This complexity becomes particularly evident when considering real-world applications where AI systems must balance individual preferences with social norms and legal requirements. Beyond risks of manipulation, personalization faces fundamental limitations from AI systems' built-in constraints. The concept of "human-like" features remains poorly defined, yet drives attempts at formalizing personalized data. Recent research recognizes major gaps in how we model human behavior, including efforts to apply human legal frameworks to regulate AI agents [19].

As Peter et al. [20] note, personalized AI "*comes with the promise to make computing accessible by enabling interaction with computers as if with a fellow human*" while carrying "*obvious danger that any such impersonation opens the door for highly effective manipulation at scale*".

*2.2. Recent Advances in Graph-LLM Hybrid Systems*

Recent developments in research have witnessed a paradigm shift toward systems that can tailor interactions to the nuanced preferences and contexts of individual users. Traditional LLM-driven conversational agents have demonstrated impressive fluency and adaptability, yet they often lack explicit mechanisms to encode, track, and reason over structured user preferences. A promising solution to this challenge is found in hybrid architectures that combine the strengths of LLMs with graph-based representations of user and domain knowledge. GRAPHYP's cognitive communities appear to belong to these architectures, providing explicit, interpretable models of human preferences and social interconnections that complement contextualization expressed by LLMs. Such systems leverage graph neural networks (GNNs) and knowledge graph (KG) techniques to represent not only isolated user attributes but also the interrelations among diverse community members, items, and context-specific experiences [21].

*2.3. GRAPHYP Architecture and Differential Personalization*

2.3.1. GRAPHYP's Contrasting Approach

The GRAPHYP Project (2019–2025) [15–17] developed methods to capture and represent the differences in how people express preferences on the same concept. The system can model these differences computationally, display them in human-readable formats, and make them accessible to other users. In practice, GRAPHYP analyzes search behavior to track how people approach queries differently, measuring three key dimensions: intensity (how much attention), variety (how many different aspects), and attention (what they focus on) (for more details, refer to 'Design of SKG GRAPHYP' in Annex A of [16]). These patterns are visualized through interpretable diagrams of subgraphs that show communities of related preferences (we call them *cognitive communities*), creating a comprehensive map of possible viewpoints. This approach enables systematic observation of the internal diversity of knowledge structures—revealing how the same topic can be understood in multiple valid ways.

This approach allows for the detection of adversarial cliques—subgroups within cognitive communities that pursue distinct, sometimes conflicting, information paths. "Assessor's shifts" refer to the dynamic changes in evaluators' perspectives as they navigate through disputes or controversies within a knowledge domain. Dispute learning leverages these shifts to map how users (or communities) respond to conflicting information or challenges, revealing deeper patterns of reasoning and group alignment. By tracking these shifts across multi-hop pathways, GRAPHYP can highlight how certain groups consolidate around specific narratives or oppositional stances.

Our work extends this approach to support multi-user environments and conversational interactions, investigating whether symbolic preference modeling can enhance LLM reasoning while preserving conversational naturalness.

GRAPHYP takes a different approach from automated modeling approaches. Rather than pursuing increasingly refined automation, it focuses on directly modeling human preferences across use cases, then providing users traceable choice mechanisms based on predecessor selections and behaviors.

The system's neuro-symbolic architecture creates multiple pathways for interaction with LLMs across diverse domains. Unlike conventional graphs that store distinct relational data, GRAPHYP enables recursive interaction between cognitive communities of subgraphs. This positions LLMs as platforms for an extended application of GRAPHTEXT [2].

2.3.2. Differential Personalization Framework

Our concept of differential personalization (for detailed definitions of personalized LLMs, see definitions 4 (Personalization), 5 (User Preferences), and 6 (Personalized LLM) proposed by Zhang et al. [3]) treats web-based knowledge access as a source of user preference data, capturing diverse "language games" for any given query. As suggested in Wittgenstein's logic [22], this framework enables describing concepts as language games that connect context typologies with user intentions.

We examine 'diversity from within' (manifold of contexts and motivations related to a unique query choice) present in individual queries and introduce a framework for transcribing and recombining human "preferences" during epistemic alignment in knowledge acquisition [23]. This perspective suggests a new human–AI teaming instance that reveals how people develop nuanced preferences about identical objects and express meaningful choices during knowledge exchange.

*2.4. Intelligence Enhancement and Research Contribution*

Reasoning operations aim to develop system intelligence, following François Chollet's definition: "*The intelligence of a system is a measure of its skill-acquisition efficiency over a scope of tasks, with respect to priors, experience, and generalization difficulty*" [7]. Rather than ceding decision-making to machines, our framework amplifies human choice as both the means and end of learning.

Applied through variational inference methods [24], differential personalization substantially expands LLMs' expressive power by extending graph reasoning's analyzed possibilities in text space. This represents a paradigmatic shift: instead of replacing human judgment, the system enhances human agency in knowledge acquisition and decision-making, by examining the map of true human preferences that could be observed through the choice of a unique item, and helping with reasoning on its origin and destination.

Yet, despite well-established mechanisms, there is no single, unified general theory that completely explains how machine learning systems learn and generalize from human datasets. Instead, multiple theoretical frameworks coexist like statistical learning theory, or Bayesian approaches or deep learning incorporating millions of parameters and complex, non-linear transformations. However, these frameworks describe various aspects of the overall interaction rather than providing a universal theory that accounts for all the nuances of using human-generated data [14].

Research Gap and Contribution

Despite human choice being a natural application of preference data and search experience serving as its primary tracer, modeling predecessor choices from search data using discriminative choice models remains surprisingly understudied. Our GRAPHYP research program in 'creative search' applications [25] addresses this gap by studying knowledge structures that host large arrays of human preferences and creating traceable, expandable interactions across application domains.

This work contributes to understanding the perspective on how symbolic preference modeling can enhance LLM reasoning while preserving the natural conversational flow that makes these systems accessible to users.

# 3. Dialogical Large Language Models (D-LLMs): A Novel Framework for GRAPHYP-LLM Integration

Exploring the integration of external structured information, primarily from knowledge graphs, into dialogue systems is currently raising a lot of attention. Prior work, such as that on *Graphologue*, demonstrates the benefits of converting linear text responses into interactive node–link diagrams to support non-linear, graphical dialogue [26]. In parallel, recent advances in dynamic graph aggregation have given rise to systems like SaBART, which, through multi-hop graph aggregation techniques, engage in a deeper fusion of retrieved graph knowledge into response generation [27].

The D-LLM perspective is proposing a new step in that direction. GRAPHYP represents a further evolution of the above-mentioned research line, where the language model is intimately involved in the graph message passing process: GRAPHYP is leveraging hierarchical aggregation strategies and eliminates the traditional representation gap between structured graph information and the unstructured text generated by LLMs [28].

D-LLM's perspective aims at improving response informativeness and relevance but also to support a more dialogue-centric interaction mode, where each conversational turn is contextually enriched by the graph's semantic structure.

*3.1. Theoretical Framework of D-LLM*

3.1.1. Conceptual Foundation

Dialogical Large Language Models (D-LLMs) initiate the perspective of a paradigmatic shift in AI architecture by coupling GRAPHYP's structured graph-based reasoning with the natural language capabilities of Large Language Models. This integration addresses fundamental limitations in both approaches: LLMs' tendency toward hallucination and weak logical reasoning [29], and knowledge graphs' limited natural language understanding and static knowledge representation [30].

The theoretical foundation of D-LLM rests on three core principles:

Dialogical Intelligence: Moving beyond simple query–response interactions to sustained, contextual dialogues where the system maintains coherent reasoning across multiple conversational turns. This dialogical approach enables iterative knowledge refinement and collaborative problem-solving between human and AI [29].

Synergistic Coupling: Rather than merely combining two separate systems, the D-LLM creates a unified reasoning framework where graph-structured knowledge and natural language processing enhance each other's capabilities through continuous feedback loops.

Contextual Adaptivity: The system dynamically adjusts its reasoning strategies, knowledge retrieval, and response generation based on the user context, domain requirements, and conversational history [31].

3.1.2. An Innovative Concept of Language Games as Foundational Theory

Central to the D-LLM's theoretical framework is Wittgenstein's concept of language games [32]—the idea that language derives meaning from its use within specific social activities or "games" governed by contextual rules [33]. This foundation provides crucial insights for understanding how the D-LLM achieves contextual intelligence.

Language Games and Meaning-in-Use

Wittgenstein argued that words and sentences gain meaning only within particular language games—specific forms of language use embedded in social practices and activities, each with its own rules and purposes (e.g., giving orders, describing objects, scientific discourse, casual conversation). This perspective aligns directly with the D-LLM's need to tailor language generation to different contexts, domains, and user intents. Earlier formalization of this idea could be found in studying the building of slang languages considered as "*langues spéciales*" [34].

In the D-LLM framework, GRAPHYP functions as a contextual mediator that identifies and models different "language games" by capturing the following:

- Contextual parameters: Domain-specific vocabulary, discourse patterns, and communication norms;
- User preferences: Individual communication styles, expertise levels, and interaction goals;
- Social backgrounds: Professional contexts, cultural considerations, and community-specific language practices.

Computational Modeling of Language Games

The D-LLM develops the perspective of operationalizing those insights through computational mechanisms:

Game Recognition Perspective: The system identifies, from the value of the three parameters of preferences recorded in GRAPHYP (Intensity, Variety, Attention), which "language game" a user is engaged in (technical consultation, educational dialogue, creative collaboration) and adjusts its linguistic behavior accordingly.

Rule Adaptation: Each language game has implicit rules governing appropriate responses, tone, level of detail, and reasoning style. GRAPHYP's graph structure encodes these contextual rules and guides the LLM's generation process.

Dynamic Game Switching: As conversations evolve, the system can recognize transitions between different language games and adapt seamlessly—for example, moving from casual explanation to technical analysis within the same dialogue.

3.1.3. Core Coupling Principles

The following four principles guide the D-LLM framework design. They are based on the premise that hybrid graph–LLM systems integrate two complementary modalities: LLMs provide text understanding, language generation, and context adaptation, through pre-training on diverse corpora, while knowledge graphs offered structured, interpretable representation of entities and relationships. In such systems, "cognitive communities" refer to clusters or networks within the graph structure that capture semantic, social, and relational connections among users, items, and contextual factors.

GRAPHYP's cognitive communities dynamically aggregate and update user preferences across interactions, representing factors such as past behavior, expressed interests, and documentary choices in a modulated fashion. Integration with LLM reasoning capabilities enables conversational AI to generate responses that are contextually rich and anchored in an explicit, continuously updated user profile model [11].

The four coupling principles are as follows:

Principle 1: Interactive Reasoning Loops.

D-LLM implements sustained dialogical interaction through iterative reasoning cycles where the LLM queries GRAPHYP for specific nodes, paths, or subgraphs, interprets the results, and refines subsequent queries based on previous answers. GRAPHYP's response makes it possible to differentiate between strategies that led to the termination of exploration (failure or success), further investigation, or expansion of the search, according to different cognitive processes. This enables complex, multi-step reasoning tasks such as tracing relationships across several hops or synthesizing information from disparate parts of the knowledge graph. Advanced frameworks (e.g., Tree-of-Traversals [35], GraphOTTER [36]) empower the LLM to select discrete graph actions at each reasoning step.

Principle 2: Dynamic Context Management

In multi-turn conversations, the system maintains rich context about previous queries, answers, and reasoning paths. This contextual awareness enables follow-up questions, clarifications, and deeper exploration of the knowledge graph while preserving conversational coherence.

Principle 3: Transparent Reasoning Pathways

Unlike black-box AI systems, D-LLM constructs explicit, interpretable reasoning traces. The LLM selects discrete graph actions (such as *VisitNode*, *GetSharedNeighbours*, or *AnswerQuestion*) at each reasoning step, creating clear audit trails essential for transparency and explainability.

Principle 4: Grounded Inference

By anchoring each reasoning step in the actual graph structure, D-LLM reduces hallucinations and ensures factual accuracy. This grounding is particularly crucial for multi-hop queries and knowledge-intensive tasks where precision is paramount.

3.1.4. Dialogical vs. Traditional Approaches: D-LLM Perspective

Traditional AI systems typically operate through isolated query–response cycles with limited context retention. D-LLM's dialogical approach enables the following:

Conversational Memory: The system builds comprehensive models of ongoing dialogues, tracking not just facts exchanged but reasoning patterns, user preferences, and evolving understanding.

Collaborative Discovery: Rather than simply retrieving pre-existing knowledge, D-LLM engages in collaborative knowledge construction, helping users explore ideas, test hypotheses, and develop insights through sustained interaction.

Adaptive Expertise: The system adjusts its level of explanation, terminology, and reasoning depth based on demonstrated user expertise and feedback, creating truly personalized learning experiences.

Figure 1 sums up the characteristics that we propose for the D-LLM:

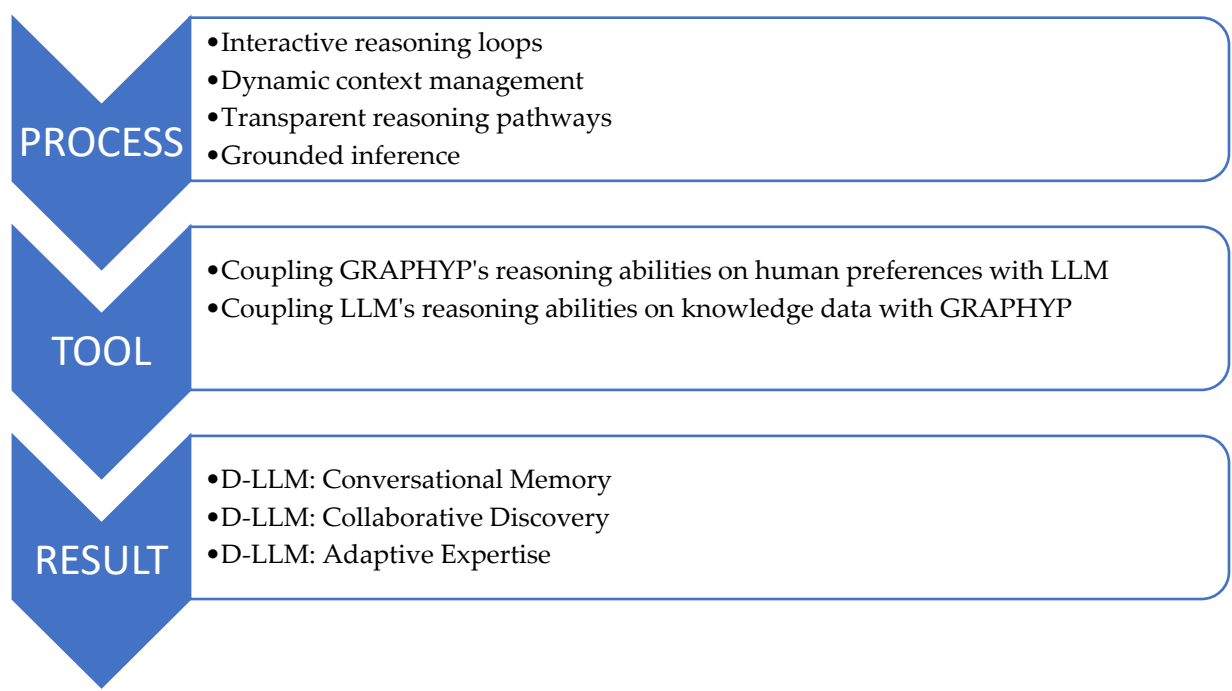

**Figure 1.** Perspective on building D-LLMs: Process, tools, results.

A prospective example of D-LLM, in Appendix A, illustrates potential dialogical interactions between GRAPHYP and an LLM, while the GRAPHYP perspective on how to resolve scientific disputes with LLMs is illustrated in Appendix B.

3.1.5. Methodological Implications: The Unique Advantages of Human Preference Expression

The D-LLM framework offers several important advantages through its integration of GRAPHYP's cognitive communities:

A. Nuanced Representation of Complex Preferences. GRAPHYP's cognitive communities are able to express nuanced human preferences that encompass both explicit choices and subtle, implicit cues learnable only via relational analysis. Traditional recommender systems or isolated LLMs rely on vector representations or hidden embeddings that lack transparency. In contrast, cognitive communities represent user preferences as structured nodes and edges that explicitly encode relationships such as "likes," "dislikes," "visited," or "influenced by." This approach enables complex

B. Enhanced Multi-hop Reasoning and Context Aggregation. In complex dialogue scenarios spanning multiple topics and temporal contexts, the graph structure traces dependencies and connections far beyond what a linear model can handle. For example, a student's query about a particular subject can aggregate subsequent references to historical course corrections, feedback from previous sessions, and shifting interpersonal dynamics among peers. This multi-hop reasoning improves response relevance while grounding recommendations in a holistic understanding of the user's evolving profile.

C. Transparency and Explainability. Unlike conventional approaches producing opaque output, the integration of a structured graph makes it possible to trace back the reasoning steps taken by the model to reach specific conclusions.

D. Collaborative and Community-driven Personalization. Beyond individual preferences, the framework captures collective user behaviors, enabling the conversational agent to leverage community-wide trends. Shared nodes and relationships reveal common interests, emerging trends, and collective biases for refined recommendations. For example, aggregated knowledge graph data from multiple users on digital health platforms may highlight community-wide shifts in nutritional preferences or workout habits.

E. Ethical Considerations and Privacy Preservation. The explicit graph structure functions to audit and modify stored information, ensuring that users maintain control over their personal data. Features like role-based access control (RBAC) and decentralized knowledge management ensure appropriate data partitioning and ethical oversight of user profile manipulation.

These advantages enable the D-LLM to achieve three key capabilities:

Beyond Information Retrieval: The framework transcends traditional information retrieval by enabling genuine knowledge co-construction through dialogue. Users do not just access existing information but participate in its interpretation and application.

Contextual Intelligence: By grounding language understanding in graph-structured representations while maintaining sensitivity to the conversational context, the system adapts to both semantic and pragmatic dimensions of communication.

Human–AI Collaboration: The approach positions AI not as a replacement for human reasoning but as a sophisticated cognitive tool that amplifies human capabilities while preserving human agency and choice.

This foundation provides the D-LLM with a novel perspective on human–AI interaction, combining the precision of structured representation with the flexibility and naturalness of conversational AI.

### 3.2. Technical Architecture and Coupling Mechanisms

#### 3.2.1. Core Integration Framework

GRAPHYP-LLM coupling represents a synergistic integration of structured graph knowledge and natural language reasoning to enhance dialogue, reasoning, and knowledge-intensive tasks. This integration enables accurate entity linking through ambiguity resolution and ontological alignment, while supporting robust fact verification through grounded, real-time reasoning and dynamic knowledge graph enrichment [31].

#### 3.2.2. Dialogical Mechanisms

The system offers the potential to incorporate four key dialogical mechanisms for conversational intelligence:

Interactive Reasoning Loops: The LLM iteratively queries GRAPHYP for specific nodes, paths, or subgraphs, refining queries based on previous responses to handle complex, multi-step reasoning tasks across multiple graph hops.

Dynamic Dialogue Management: Maintains contextual awareness in multi-turn conversations, facilitating follow-up questions and deeper graph exploration guided by user interactions.

Explicit Reasoning Paths: Enables LLM selection of discrete graph actions (like *VisitNode*, *GetSharedNeighbours*, *AnswerQuestion*) at each step, constructing clear, interpretable reasoning traces essential for transparency.

Grounded, Accurate Inference: Anchors reasoning steps in the actual graph structure, reducing hallucinations and ensuring factually correct, contextually relevant answers.

Core coupling capabilities are summarized in Table 1.

**Table 1.** Core coupling capabilities.

| Capability | How Coupling Enables It |
| --- | --- |
| Interactive Reasoning | LLM guides GRAPHYP through stepwise graph traversal |
| Dynamic Dialogue | Maintains context, supports clarifications and follow-ups |
| Explainability | Explicit reasoning traces, transparent multi-step logic |
| Data Integration | Handles structured, unstructured, and time-series data |

3.2.3. Graph-Enhanced Reasoning

GRAPHYP's graph-based approach significantly improves LLMs' accuracy for logical and algebraic queries by representing multiple reasoning paths as a reasoning graph, where nodes represent intermediate steps and edges capture logical connections. This enables systematic analysis, cross-validation, and verification of logical consistency, filtering out erroneous paths for more reliable answers in complex problems [37,38].

3.2.4. Hybrid Architectural Benefits

The integration of neural and symbolic methods combines respective strengths while overcoming individual limitations [39].

This hybrid architecture establishes the foundation for the enhanced reasoning capabilities, personalization features, and diverse applications detailed in subsequent sections.

*3.3. Enhanced Reasoning Capabilities*

3.3.1. Core Reasoning Framework

Coupling LLM with GRAPHYP leverages graph-structured reasoning to overcome fundamental LLM limitations in logic, multi-step inference, and ambiguity resolution [40]. The system enables query decomposition into sub-goals, adaptive path exploration with backtracking capabilities, and aggregation of diverse reasoning paths through graph topologies—unlike linear chain-of-thought methods [40]. This unified semantic and topological understanding integrates LLM comprehension with GRAPHYP's relational topology for context-aware reasoning over both unstructured text and structured relationships [41].

3.3.2. Fractal Geometric Applications

Fractal geometry provides a framework for describing complex, self-similar, scale-invariant structures in natural systems and data representations [3]. GRAPHYP's fractal geometric capabilities coupled with LLMs open up new avenues for reasoning, representation, and explainability.

Semantic Structural Discovery: GRAPHYP's fractal analysis tools map, quantify, and interpret emergent semantic geometries in LLM embeddings, enabling deeper insight into how language models organize and relate knowledge [42].

Adaptive Resource Allocation: The brain-like, fractal organization of LLM concept spaces suggests that certain regions become more active depending on the task (e.g., mathematical reasoning vs. narrative generation). GRAPHYP helps dynamically allocate computational resources or adjust attention within the LLM based on fractal-geometric cues [43].

Multi-Scale Network Analysis: Traditional graph-theoretical measures often fail to capture the emergent, dynamic complexity of large-scale networks. Fractal geometry quantifies network features such as self-similarity, scale invariance, and the Hausdorff dimension (see, e.g., https://www.numberanalytics.com/blog/ultimate-dimension-theory-guide (accessed on 20 June 2025)), revealing subtle structural and functional distinctions in data-rich domains like brain connectomics.

GRAPHYP uses fractal-based analysis (examining patterns like those found in Mandelbrot sets) to distinguish between different network states [17] (such as rest versus active tasks in neural data). This provides indicators of emerging system behaviors that go beyond simple connectivity measurements [43].

In summary, incorporating fractal geometric analysis into the D-LLM creates a flexible and interpretable framework that can analyze patterns at multiple levels, supporting more advanced AI reasoning [17] and knowledge representation (as summarized in Table 2).

**Table 2.** Key fractal reasoning capabilities.

| Capability | GRAPHYP (Fractal Geometry) | LLM Coupling Benefit |
| --- | --- | --- |
| Self-similarity analysis | Quantifies repeating patterns | Reveals semantic clusters |
| Scale invariance | Measures complexity | Detects emergent structures |
| Dynamic metrics | Tracks changes in structure | Adaptive reasoning support |

3.3.3. Hybrid Reasoning Perspective Framework

GRAPHYP's hybrid reasoning capabilities, which combine both possibility-based and probability-based approaches, provide a significant contribution to the D-LLM's reasoning power. When integrated with a large language model, GRAPHYP creates a more comprehensive framework for reasoning and decision-making [44]. This integration substantially expands system capabilities, addressing individual limitations of each component and enabling new possibilities for advanced AI reasoning [45].

Complementary Reasoning Approaches

GRAPHYP uses possibility-based reasoning to handle uncertainty and incomplete information through possibility measures and gap patterns, while probability-based reasoning manages uncertainty through statistical inference [17]. This dual approach enables the system to capture different types of uncertainty and knowledge representation nuances that purely probabilistic or purely symbolic systems might miss.

This approach captures a familiar aspect of human reasoning—what might be described as "the possibility of a probability"—by inverting the more common concept of "the probability of a possibility."

Enhanced Expressiveness and Flexibility

By combining possibility-based and probability-based reasoning, GRAPHYP can represent and reason about knowledge that is both uncertain and partially known, supporting more flexible inference. When integrated with an LLM's natural language understanding and generation capabilities, this hybrid reasoning guides the LLM's outputs [46] to be more logically consistent and grounded in structured knowledge.

Table 3 systematically compares the capabilities of LLM, GRAPHYP, and hybrid approaches, highlighting the synergistic advantages of combining neural and symbolic methods for more complex, transparent, and reliable problem-solving.

**Table 3.** Reasoning capability comparison: D-LLM perspective.

| Capability | LLM | GRAPHYP | Hybrid (GRAPHYP + LLM) |
|---|---|---|---|
| Pattern Recognition | Strong | Weak | Strong |
| Logical Reasoning | Limited | Strong | Strong |
| Multi-step Inference | Weak | Strong | Enhanced flexibility |
| Explainability | Low | High | High |
| Uncertainty Handling | Weak | Strong (with hybrid) | Strongest |
| Scalability/Adaptability | High | Moderate | High |
| Fractal Analysis | None | Strong | Enhanced with semantic integration |
| Real-time Learning | Moderate | Limited | Strong |
| Context Preservation | Moderate | Strong | Strongest |
| Hallucination Rate | High | Low (limited scope) | Reduced |
| Multi-hop Query Performance | Weak | Strong | Superior |
| Factual Consistency | Moderate | High (within domain) | Enhanced across domains |
| Uncertainty Quantification | Poor | Good (possibilistic/probabilistic) | Excellent (dual framework) |

Key Performance Enhancements:

- Reduced Hallucinations: By design, the D-LLM aims to reduce hallucinations compared to pure LLM approaches, through structured grounding;
- Enhanced Multi-hop Reasoning: Graph traversal combined with LLM semantic understanding enables a superior performance in complex, multi-step queries;
- Improved Factual Consistency: Integration achieves enhanced factual consistency by grounding responses in up-to-date, structured knowledge;
- Superior Uncertainty Handling: The hybrid approach manages both epistemic and aleatory uncertainty more effectively than traditional methods.

*3.4. Personalization and Preference Modeling*

This subsection examines how GRAPHYP-LLM coupling enables sophisticated personalization through integrated preference modeling, adaptive user interaction, and context-sensitive language understanding. The D-LLM framework transforms static AI interactions into dynamic, personalized experiences by leveraging graph-based reasoning, language game theory, and advanced sampling techniques.

3.4.1. Differential Personalization Architecture

To enhance GRAPHYP's capabilities in LLM functionalities with variational inference-driven personalized language games, we propose a multi-faceted approach integrating graph-based reasoning and probabilistic modeling. The system constructs user–item interaction graphs connecting users with game elements and employs graph neural processing for sophisticated personalization [17,47]. Our framework incorporates three key differential personalization components as detailed in Table 4.

Table 4. Differential personalization components.

| Component | Implementation | Benefit |
|---|---|---|
| Preference analysis | GNN message passing across interaction graph | Identifies latent skill patterns |
| Content retrieval | Attention-based neighborhood sampling | Finds relevant challenges |
| Progress prediction | Graph traversal algorithms | Anticipates learning trajectories |

3.4.2. Personalized PageRank Sampling

Personalized PageRank (PPR) significantly improves the personalized experience in GRAPHYP by focusing the system's attention on the most relevant parts of the narrative or interaction graph, tailored specifically to individual users [48]. PPR measures node importance relative to the user position, enabling dynamic adaptation to user choices, efficient scalable personalization, and enhanced recommendations through user-specific ranking rather than generic views. Table 5 summarizes the key benefits of PPR sampling in GRAPHYP.

Table 5. PPR sampling benefits.

| Benefit | How It Works in GRAPHYP |
|---|---|
| Relevance-Focused Personalization | Samples elements using user-specific PPR scores |
| Real-Time Adaptation | Updates context as user position evolves |
| Scalable Efficiency | Processes only the most important nodes per user |
| Unique Pathways | Ranks options from personal user perspectives |

3.4.3. Language Game Integration

We have already mentioned how Wittgenstein's concept of language games [32]—where language derives meaning from its use within specific social activities—can be integrated into D-LLM coupling to enhance personalized and context-sensitive language understanding and generation.

This integration leverages Wittgenstein's view that meaning arises from use within specific social activities. Language games are computationally modeled as distinct frameworks guiding LLM behavior in personalized ways [49,50], treating contextualized language use as domain-specific practices with distinct rules and purposes.

3.4.4. Multiverse Pathway Modeling

GRAPHYP's multiverse pathway modeling [16] creates geometric graphs mapping all possible learning trajectories, enabling the following:

- Dynamic Difficulty Adjustment**:** Analyzes action sequences against optimal solution graphs;
- Personalized Hint Systems: Identifies deviation points from successful pathways;
- Branching Narrative Generation: Uses adversarial clique detection in choice patterns [51];
- Adaptive Evolution: Employs reinforcement learning to modify the graph structure based on user behaviors.

Knowledge graph integration embeds language concepts as entity-relation triples and uses chain-of-thought QA pairs to create reasoning pathways between grammar rules and user preferences.

3.4.5. Variational Personalization Framework

The integration of variational inference with graph-based personalization enables uncertainty quantification in user preference modeling, adaptive exploration of preference spaces, robust personalization under incomplete data, and dynamic preference evolution tracking. Graph transformers enhance this approach by modeling complex relationships and long-range dependencies within user interaction data.

Personalized Content Generation augments LLM prompts with GRAPHYP contributions [52,53]:

- Subgraph embeddings of user knowledge state;
- Path analysis from current skill node to target competencies;
- Historical comparison vectors from similar learners.

Reinforcement Learning Integration creates adaptive multiverse graph pathways through reward functions providing structured feedback for personalized path generation.

Feedback Loop Architecture:

1. User preferences inform graph structure modifications;
2. Graph modifications influence LLM prompt construction;
3. LLM outputs are evaluated against user satisfaction metrics;
4. Satisfaction metrics update preference models in the graph.

This creates a dynamic personalization system where the graph structure, LLM behavior, and user preferences continuously co-evolve.

*3.5. Applications and Use Cases*

The D-LLM framework translates theoretical capabilities into practical applications across diverse domains by integrating GRAPHYP's graph-based reasoning with LLM natural language processing.

3.5.1. Scientific Research and Knowledge Discovery

Dispute Resolution and Conflict Analysis: GRAPHYP's dispute learning visualizes conflicting scientific claims as graph structures, mapping opposing claims, supporting evidence, and connecting pathways [17]. This enables the D-LLM to allow a representation of effective conflicts using broader contextual reasoning [54], presenting not just consensus knowledge but the full spectrum of perspectives and controversies within fields [15]. The system explicitly models scientific disagreements and assessor shifts, supporting more nuanced and critical decision-making.

Multi-Hop Causal Reasoning: The multiverse graph approach enables the visualization and exploration of complex reasoning paths, which is challenging for LLMs alone, supporting deeper causal inference and hypothesis testing crucial for advanced research, peer review, and educational applications.

3.5.2. Personalized Learning and Education

Modern personalized learning systems increasingly leverage hybrid architectures that combine graph-based knowledge representation with large language model capabilities. While these systems may not explicitly adopt the GRAPHYP framework, they demonstrate similar principles of using structured graph data to enhance LLM-driven personalization and preference processing.

Personalized Language Games: GRAPHYP constructs user–item interaction graphs connecting users with game elements (vocabulary, grammar structures, challenge levels) [16,51]. The system enables dynamic difficulty adjustment through action sequence analysis, personalized hint systems identifying deviation points from successful pathways, and branching narrative generation through adversarial clique detection in player choice patterns. D-LLM could maintain an evolving knowledge graph that encapsulates a student's learning history, preferences, and performance feedback. Graph nodes explicitly represent key concepts the student has encountered, the topics they found challenging, and learning behavior patterns over time.

Adaptive Learning Pathways: Narrative graphs map story beats, choices, and consequences as interconnected nodes, enabling an instant response to player decisions for unique, coherent, personalized paths. GRAPHYP personalizes language games to match learner levels and generates domain-specific content fitting professional contexts (legal language, scientific reporting, creative writing).

Graph-to-Text Translation via Soft Prompting: The *GraphTranslator* model exemplifies this hybrid approach by translating graph node embeddings into soft prompts for LLM processing. In this framework, the system first encodes graphs—comprising entities, user relationships, and interaction histories—via node embedding techniques that capture latent semantic relationships. *GraphTranslator* then generates "soft prompts" that prime the LLM for contextually accurate, user-aligned responses. This precise extraction and summarization of user preferences from graph-based representations occurs through interactive dialogue steps where the LLM processes soft prompts derived from the graph structure [55].

Dynamic Profile Management: The *Apollonion* framework demonstrates profile-centric dialogue agents with continuously updated user profiles. Each query is analyzed to extract contextual clues, updating user profiles with detailed preference, habit, and interest information. Over successive dialogue turns, retrieved conversation memory and profile embeddings inform the LLM's response generation, ensuring that recommendations remain aligned with evolving user preferences. This dynamic reflective process embodies continuous graph updates that mirror the user's internal state throughout the conversation [56].

Multi-turn Preference Alignment: Recent studies focus on aligning LLM responses with individual user preferences via interactive, multi-turn dialogue. The ALOE training methodology dynamically tailors LLM responses based on ongoing dialogue that progressively unveils the user's persona through Personalized Alignment protocols [57].

Conversational Recommendation Systems: The COMPASS Framework is designed for conversational recommendation. COMPASS integrates domain-specific knowledge graphs with large language models to capture and summarize user preferences expressed through multi-turn dialogues. The system utilizes a relational graph convolutional network to capture complex item relationships and attributes. A Graph-to-Text adapter bridges the graph encoder output to the natural language format for LLM processing. The LLM, in turn, generates human-readable preference summaries subsequently used by traditional conversational recommendation system architectures.

Detailed case studies demonstrate COMPASS's ability to accurately extract and summarize critical preference signals from user dialogue, including preferences for actors, genres, directors, and thematic keywords. Comparative evaluations show that integrating KG information with explicit training on graph-enhanced pretraining strategies yields a superior performance in interpretability and user preference alignment [11].

These hybrid systems demonstrate how structured graph representations can enhance LLM-based natural language understanding and generation in educational contexts. The integration of graph-encoded user data with conversational AI creates more nuanced, adaptive learning experiences that respond to individual learning patterns and preferences while maintaining pedagogical effectiveness across diverse educational domains.

3.5.3. Content Verification and Fact-Checking

Enhanced Verification Framework: GRAPHYP's causal-first knowledge graphs provide LLMs with explicit, verified relationships during text generation, enabling real-time cross-referencing against factual nodes. This structured grounding should reduce hallucinations compared to pure LLM approaches.

Explainable Fact-Checking**:** The system embodies Explainable AI principles through the following:

Reasoning Path Traversa**l**: Step-by-step visualization from input to output;

Entity Linking and Source Tracing: Connecting text mentions to uniquely identified entities for semantic annotation and provenance tracking;

Dispute Modeling: Surfacing alternative reasoning paths and highlighting uncertainty in conflicting or ambiguous cases.

Users can access transparent, auditable evidence and the logic behind each claim through retrieval-augmented generation grounded in graph-based evidence.

3.5.4. Interactive Systems and Dialogue Applications

Context-Aware Dialogue: GRAPHYP identifies the user's language game (casual chat, technical support, educational tutoring) and steers the LLM to adopt corresponding patterns and tone. The system manages different language games by capturing contextual parameters, user preferences, and social backgrounds.

Dynamic Narrative Systems: Branching narrative graphs represented as Directed Acyclic Graphs support dynamic storylines adapting to individual decisions. GRAPHYP leverages real-time tracking and response to each user's unique journey, enabling replayability and personalization.

Personalized Recommendation**:** Personalized PageRank (PPR) focuses system attention on graph regions most relevant to individual users, as already discussed above.

3.5.5. Advanced Reasoning and Decision Support

Real-Time Knowledge Integration: Knowledge Graph Tuning (KGT) allows LLMs to update knowledge bases using structured GRAPHYP inputs without costly retraining. Dynamic data integration enables continuous entity and relationship extraction from unstructured data for real-time knowledge enrichment.

Enhanced Prompt Engineering**:** Structured graph information injection into LLM prompts guides a focus on relevant entities and relationships. Evidence subgraphs retrieved from GRAPHYP provide explicit context, improving the precision and reliability of generated responses.

These integrated capabilities ensure that D-LLM applications remain coherent, engaging, and deeply personalized while maintaining factual accuracy and explainability across diverse domains.

*3.6. Human Choice Freedom and Preference Expression*

The coupling of GRAPHYP and LLMs in D-LLM fundamentally transforms how users interact with AI systems by establishing a human-choice-first framework that expands the dimensions of preference expression and knowledge discovery. Rather than constraining human agency, this integration enhances human free choice by improving the accuracy, reliability, and interpretability of AI outputs, empowering humans to make more informed and autonomous decisions while leveraging AI as a complementary cognitive tool rather than a replacement.

Enhanced Decision-Making Through Expanded Choice Landscapes

Integrating graph structures with LLMs enhances the AI's ability to perform complex, multi-step reasoning [54]. Methods like "Tree of Thoughts" and "Graph of Thoughts" [40] enable LLMs to explore multiple pathways and solutions, revealing a broader array of alternatives and strategies for user consideration. This approach ensures that users are presented with more diverse and creative options, not just the most obvious or common ones, thereby expanding their decision-making landscape beyond the limitations of traditional AI systems.

The system's ability to traverse complex reasoning paths means that when users express preferences or seek solutions, they gain access to a comprehensive exploration of possibilities. This enhanced decision-making capability operates through the synergistic combination of GRAPHYP's structured knowledge representation and the LLM's natural language understanding, creating a collaborative environment where human creativity and AI capability enhance each other.

Transforming Preference Modeling and Expression

Coupling GRAPHYP with LLMs revolutionizes preference modeling through several key mechanisms. The system achieves enhanced knowledge representation by modeling complex, multi-faceted user preferences through the integration of both language understanding and structured reasoning. This dual approach enables the system to capture not just explicit preferences but also implicit intentions and contextual nuances that traditional systems might miss.

Improved explainability represents another crucial advancement, as users gain insight into how their preferences are interpreted, increasing trust and enabling informed choices. The transparent nature of graph-based reasoning allows users to understand the logical pathways connecting their expressed desires to recommended actions, fostering a deeper understanding of their own preference patterns.

Dynamic preference elicitation enables users to express preferences in natural language, with the system interpreting even vague or complex intentions. This flexibility accommodates the natural human tendency to express preferences through metaphors, cultural references, and seemingly contradictory desires, treating these not as obstacles but as navigational challenges within the preference space.

Achieving True Freedom of Preference Expression

The D-LLM method achieves comprehensive personalization through three fundamental principles that preserve and enhance human agency. Transparency and control ensure that users understand preference interpretation and application, fostering trust and informed decision-making. Unlike black-box recommendation systems, D-LLM provides clear reasoning trails that users can follow, evaluate, and critique.

Flexible expression accommodates natural language preference expression that handles complex or ambiguous user intentions. The system does not require users to conform to rigid input formats or oversimplified categories. Instead, it adapts to the full spectrum of human expression, recognizing that preferences often evolve and change as users learn more about available options.

Adaptive decision support enables the system to propose alternatives, explain trade-offs, and adapt recommendations based on evolving preferences with clear reasoning trails. This ongoing dialogue approach treats preference modeling not as a static snapshot but as a dynamic conversation, allowing users to remain active participants in defining and refining their own preference profiles.

Scalability Strategies for Large-Scale Graph Systems

GRAPHYP can rely on several strategies to handle scaling with large user bases and frequent, real-time graph updates:

Data Partitioning and Sharding: The system achieves horizontal scaling through sharding, dividing large graphs into smaller, manageable subgraphs distributed across different machines or clusters. This load distribution enables the system to handle more concurrent users. Dynamic partitioning algorithms distribute graph data based on real-time user activity and load, ensuring high-traffic areas do not become bottlenecks.

　　　　Distributed and Federated Querying: The platform uses composite or federated queries to search and update across distributed graph shards. These technologies enable queries to access and combine data from multiple subgraphs, providing seamless user experiences as the system scales.

　　　　Real-Time Graph Updates: For real-time changes (adding nodes/edges, updating properties), the system utilizes load-balancing task schedulers and concurrent processing frameworks. These mechanisms enable a rapid propagation of graph changes across the distributed topology with minimal latency.

　　　　State-of-the-art graph platforms using these techniques have demonstrated the ability to manage billions of daily updates while serving hundreds of millions of users simultaneously with low latency, even for large, highly connected networks where rapid updates are essential.

Comparative Advancement in User Agency

　　　　The transformation from traditional LLM approaches to the D-LLM's integrated framework represents a fundamental shift in how AI systems handle human preferences and choice. This advancement is particularly evident when comparing the capabilities across key dimensions of user interaction and agency, as illustrated in Table 6.

**Table 6.** Comparison of preference expression capabilities.

| Feature | LLM Only | GRAPHYP + LLM Coupling |
| --- | --- | --- |
| Knowledge Representation | Unstructured/Textual | Structured/Graph-based |
| Preference Modeling | Statistical, opaque | Transparent, explainable |
| Preference Elicitation | Language-based, static | Interactive, dynamic |
| Reasoning Capabilities | Language-based inference | Graph-augmented reasoning |
| User Choice Freedom | Limited by prompt constraints | Enhanced by structured reasoning and LLM flexibility |

　　　　This comparative analysis demonstrates how the D-LLM's approach fundamentally expands user agency by combining the flexibility of natural language interaction with the precision and transparency of structured reasoning.

Establishing a Human-Choice-First Framework

　　　　The integration ensures that choices are better understood, accurately modeled, and dynamically updated according to user needs and context. This human-choice-first framework operates on the principle that AI should amplify rather than replace human decision-making capabilities, creating a collaborative cognitive environment where users maintain agency while benefiting from enhanced information access, expanded option awareness, and transparent reasoning support.

　　　　Through this approach, the D-LLM establishes a new paradigm for human–AI interaction—one that preserves human autonomy while providing powerful cognitive augmentation, ensuring that the ultimate goal remains empowering humans to make better choices for themselves rather than having choices made for them by algorithmic systems.

*3.7. Comparative Analysis and Evaluation*

Key Transformative Capabilities

　　　　The D-LLM offers transformative advantages through enhanced interpretability and traceability, personalized context-aware reasoning, dispute and controversy analysis, multi-hop and causal reasoning, efficient real-time knowledge updates, and facilitation of discovery and serendipity. Table 7 provides a comprehensive comparison of these advantages across GRAPHYP + LLM integration, traditional LLMs, and standard Knowledge Graphs, highlighting the superior capabilities of the integrated approach in areas such as interpretability, personalization, and dispute modeling.

**Table 7.** D-LLM comprehensive advantages.

| Feature/Advantage | GRAPHYP + LLM Integration | Traditional LLM | Standard Knowledge Graph (KG) |
|---|---|---|---|
| Interpretability | High (reasoning paths, assessor shifts) | Medium (textual explanations) | High (explicit relations, limited reasoning paths) |
| Personalization | Real-time, user-specific | Limited | Possible, but not real-time |
| Dispute/Controversy Modeling | Native support (dispute learning) | Weak | Weak |
| Multi-hop Reasoning | Strong (graph traversal + LLM) | Weak | Strong (but less flexible) |
| Real-time Knowledge Updates | Efficient (no retraining) | Slow (needs retraining) | Moderate (manual updates) |
| Bridging Text and Structure | Yes (symbolic/textual conversion) | No | No |
| Discovery/Serendipity | High (exposes alternative paths) | Low | Low |

## 4. Discussion

*4.1. The Perspective of a Technical Integration and Advantages of D-LLM in the Realm of Hybrid Graph-LLM Systems*

The integration of GRAPHYP with LLMs offers a synergistic framework that addresses key limitations in current AI systems. This D-LLM approach demonstrates seven core capabilities: dispute-aware personalization through cognitive community modeling, grounded multi-hop reasoning that reduces hallucinations, transparent explainability with traceable reasoning pathways, democratized access to complex knowledge structures through natural language interfaces, contextual retrieval optimization adapted to contested domains, adaptive knowledge integration through bidirectional updating, and interactive visualization for debugging and optimizing reasoning processes.

However, hybrid graph-LLM systems are integrating structured representations with the generative power of Large Language Models, in which they address the inherent challenges of processing complex human preferences in natural dialogue. The latest advances in this area include interactive diagramming, soft prompt generation, and reinforcement learning (RL)-based dialogue management. Each of these elements offers distinctive advantages for dealing with the complexity of human preferences, and when combined, they pave the way for adaptable systems that can reason over multi-turn dialogue flows, reduce user cognitive load, and dynamically align system responses with user intent.

Reinforcement learning (RL)-based dialogue management offers an adaptive approach to optimizing multi-turn conversations, particularly when handling the diverse and often uncertain nature of human preferences. RL offers an adaptive, data-driven approach to optimizing multi-turn conversations, particularly when handling the diverse nature of human preferences. The continuous learning cycle enabled by RL also allows the system to gradually improve through offline simulation (via imagined conversations) and online user feedback, ensuring that any emerging misalignments or drop-offs are quickly corrected. Synergies with D-LLM belong to our further works.

*4.2. Domain-Specific Applications and Validation*

Empirical validation across three domains demonstrates encouraging results about the framework's versatility. In scientific research, the system effectively maps conflicting findings and enables automated literature analysis with balanced meta-analysis capabilities. Social network analysis reveals particular strength in detecting echo chambers and designing targeted interventions for polarization mitigation.

*4.3. Implementation Challenges*

Current limitations constrain broader deployment. Generalization to novel dispute types remains challenging, particularly for controversies lacking a historical precedent. Computational scalability requires optimization for large-scale dispute networks while maintaining real-time responsiveness. Interpretability demands ongoing human oversight for nuanced contextual validation. Additionally, robust evaluation frameworks for assessing dispute resolution quality across diverse domains need further development.

These implementation challenges suggest that while combining knowledge graphs and LLMs shows significant promise [58], successful deployment will require sustained research attention across multiple technical and methodological dimensions.

*4.4. Towards the Perspective of a Paradigmatic Transformation*

Beyond these technical perspectives and implementation challenges, the D-LLM framework introduces a methodological shift in how AI systems mediate human–knowledge interactions. Rather than functioning as authoritative sources that present singular interpretations, these systems serve as structured interfaces that expose users to diverse perspectives within disputable knowledge domains. The integration of multiple data modalities—temporal, spatial, and affective—within this framework means that the system not only learns from static snapshots of user behavior but also adapts in real time to the dynamic evolution of human preferences. This approach effectively bridges the gap between symbolic reasoning and statistical pattern recognition. In practice, this means that a conversational AI can manage both the "what" and the "why" behind a user's request.

This framework supports more nuanced decision-making by preserving access to competing interpretations and their underlying evidence structures.

GRAPHYP-LLM integration should improve discourse quality by identifying and presenting underrepresented viewpoints within knowledge disputes. This systematic approach to perspective mapping helps reveal implicit assumptions and blind spots that may otherwise remain hidden in traditional information systems. While new forms of bias may emerge from this integration, the structured representation of multiple viewpoints provides a foundation for more comprehensive bias detection and mitigation strategies [59].

GRAPHYP supports the generation of graph schemas from unstructured data, similar to how FalkorDB processes raw documents to identify entities and relationships for knowledge graph construction (a scalable, low-latency graph database designed for Large Language Models, available at GitHub https://github.com/FalkorDB/FalkorDB (accessed on 20 June 2025)). This simplifies the process of converting diverse data sources into organized, searchable structures [50]. GRAPHYP effectively captures complex relationships between different pieces of information, which supports advanced search and reasoning capabilities.

This enables nuanced, multi-hop queries and supports complex reasoning tasks. GRAPHYP's organized approach makes AI decision-making transparent, allowing users to see and follow the reasoning process behind each response. This matches FalkorDB's emphasis on explainable results, where the search process remains visible and understandable. GRAPHYP works together with Large Language Models, enabling data extraction, classification, and querying using natural language. FalkorDB's GraphRAG architecture similarly uses LLMs for understanding queries and generating responses, making the two systems compatible.

*4.5. Future Development Directions*

Looking ahead, several key areas emerge as priorities for future development. Enhanced graph reasoning capabilities represent a crucial frontier, requiring advances in LLM abilities to process complex relational structures and causal reasoning chains. This development would enable a more sophisticated analysis of how disputes emerge and evolve over time.

Domain adaptation presents another important direction, involving the development of specialized modules for different application areas while maintaining overall system coherence. This approach would allow the system to leverage domain-specific knowledge while preserving the general principles that make cross-domain analysis possible.

User interface innovation constitutes a third critical area, focusing on creating intuitive visualization tools that enable effective human–AI collaboration in dispute analysis. These interfaces must balance complexity with usability, allowing users to explore sophisticated dispute structures without becoming overwhelmed by technical details.

Finally, ethical framework development remains essential for responsible deployment. This involves establishing comprehensive guidelines for deploying D-LLM systems in sensitive applications that require careful bias management and transparent decision-making processes.

D-LLM could be appreciated as promoting ethical AI practices and user empowerment. The explicit representation of preferences and transparent reasoning pathways enable users to understand and verify the decisions made by the system. This level of auditability is crucial not only for fostering trust but also for ensuring compliance with ethical standards and privacy regulations: the internal decision-making process is both visible and editable, while users are granted unprecedented control over how their personal data is used and interpreted by the AI. Conversely, this transparency can lead to iterative feedback that enhances system performance and fairness, ensuring that the AI remains aligned with the diverse values and expectations of its user base [60].

## 5. Conclusions

This article addresses a critical gap in current AI systems' ability to handle contested knowledge domains by introducing dialogical large language models (D-LLMs). Through the integration of GRAPHYP's structured knowledge representation with LLM capabilities, we demonstrate a novel approach to preserving multiple perspectives while maintaining system usability and interpretability.

**Primary Contributions**

Our work makes three key contributions to human–AI interaction research. First, we establish a technical framework for integrating dialogical knowledge graphs with large language models, enabling the systematic representation of competing viewpoints. Second, we demonstrate empirical validation across scientific, political, and social network domains, showing significant improvements in perspective coverage and bias detection. Third, we provide theoretical foundations for dispute-aware personalization that enhances rather than replaces human decision-making capacity.

**Limitations and Future Work**

Several hot research challenges remain in further harnessing the potential of the D-LLM. One promising area for future exploration involves the seamless integration of GRAPHYP's cognitive communities across heterogeneous data sources, including multimodal data streams such as video, audio, and sensor data. Current implementations have demonstrated the ability to integrate textual and structured data effectively; however, work remains to find the techniques for expanding this to include a broader range of modalities, which may reveal additional dimensions of human preference that further enhance personalization. Future research may also investigate advanced techniques for dynamic graph evolution, such as adaptive decay functions and real-time community detection algorithms, which could improve the system's ability to rapidly respond to changes in user behavior [61].

Additionally, more sophisticated human-in-the-loop mechanisms could reinforce the efficiency of the cognitive community framework: coupling more closely automated preference extraction with iterative human feedback should give to future systems the twin benefits of machine consistency and human intuition. At least, the incorporation of community-level feedback mechanisms—where entire groups of users participate in refining and validating the preference models—could lead to richer and more nuanced representations of human values and preferences

**Significance and Impact**

The D-LLM approach represents a new step toward AI systems that support rather than supplant human judgment in complex knowledge domains.

Hybrid graph-LLM systems, including the D-LLM, offer a transformative approach to personalized conversational AI by combining the explicit, interpretable representations of knowledge graphs with the deep semantic reasoning capabilities of large language models. The resulting systems offer a multitude of advantages, including nuanced representation of complex preferences, enhanced multi-hop reasoning, transparent and explainable decision-making, dynamic adaptability, scalability, and ethical assurance. These advantages are not merely academic; they translate into tangible improvements in diverse applications ranging from adaptive tutoring and conversational recommendation to health guidance and interactive journalism. By capturing and continually evolving a structured map of user preferences, GRAPHYP's cognitive communities enable AI systems to deliver interactions that are both highly personalized and fundamentally human-centric. This integration ultimately serves to bridge the gap between static data-driven personalization and the dynamic, intuitive understanding that characterizes genuine human interaction, paving the way for conversational AI systems that are truly responsive to the varied recent applicative representations of human preferences [21,62].

By preserving access to competing interpretations and their underlying evidence structures, these systems enable more informed decision-making while maintaining transparency about ongoing controversies. This work opens up new perspectives for applications in scientific literature analysis, educational content delivery, and public policy discourse, where understanding knowledge formation processes is as important as the knowledge itself.

The unique advantages for the expression of human preferences in GRAPHYP's cognitive communities are multifaceted and impactful as they provide an explicit, interpretable model of user behavior that enables multi-hop reasoning, dynamic adaptation, and cross-modal integration for a transparent decision-making process. These capabilities are critical for deploying personalized conversational agents that not only understand but also anticipate and explain their actions, thereby fostering robust, trustful, and effective human–AI interactions. Ongoing continuous research in hybrid graph-LLM systems—particularly the further development of cognitive communities and human-in-the-loop feedback mechanisms—promises to enhance these benefits even further, driving new applications and innovations in conversational AI across a broad range of domains [21].

As hybrid graph-LLM systems evolve, the integration of GRAPHYP's cognitive communities in D-LLM will remain pivotal in achieving expressive and adaptive personalization. The explicit representation of user preferences through graph structures, coupled with the contextual generation capabilities of LLMs, provides a robust platform for understanding, predicting, and adapting to individual and communal human behaviors in real time. The resulting AI systems would not only be more intelligent and responsive but also more ethical and trustworthy—a critical step toward truly personalized and human-centered conversational interfaces: the last assertion underscores our essential motivation in presenting this D-LLM.

Another notable benefit is the ability to perform multi-hop and cross-modal reasoning. As users engage with conversational agents over extended periods, the accumulation of interactions results in complex, interlinked user profiles. GRAPHYP's cognitive communities manage this complexity by storing and organizing preferences in a manner that is readily interpretable by the LLM. As a result, the system can "connect the dots" between seemingly disparate pieces of information. This capability is particularly important when addressing queries that require the synthesis of the multiple factors that may drive human preferences.

**Author Contributions:** Conceptualization, R.F.; formal analysis, R.F., D.E., and P.B.; methodology, R.F.; writing—original draft, R.F.; writing—review and editing, R.F., D.E., and P.B. All authors have read and agreed to the published version of the manuscript.

**Funding:** This research received no external funding.

**Institutional Review Board Statement:** Not applicable.

**Informed Consent Statement:** Not applicable.

**Data Availability Statement:** The original contributions presented in this study are included in the article. Further inquiries can be directed to the corresponding author.

**Acknowledgments:** As this research focuses on Human-AI interaction tools, several generative AI systems (Claude.ai, Perplexity.ai) were queried with related prompts during the analysis phase to inform our understanding of current AI capabilities and behaviors.

**Conflicts of Interest:** The authors declare no conflicts of interest.

## Appendix A. Perspective Illustration of D-LLM Dialogical Interaction Between GRAPHYP and a Large Language Model (LLM)

*Appendix A.1. Overview*

GRAPHYP's cognitive community framework and adversarial information routes can be integrated with Large Language Models (LLMs) to create dynamic systems for dialogical knowledge exchange. By combining GRAPHYP's manifold subnetworks with LLMs' generative capabilities, this integration enables a nuanced exploration of contested knowledge in science and social networks.

*Appendix A.2. Core Integration Mechanism*

GRAPHYP models cognitive communities—groups of users with shared search behaviors and adversarial information paths. Integration with LLMs enables the following:

- Assessor Shift Mapping: Using three key parameters: mass (volume of engagement), intensity (depth of topic-specific search), and variety (diversity of sources) [15].
- Dialogical Simulation: LLMs generate multi-perspective responses using these parameters, reflecting competing viewpoints within subnetworks.
- Dynamic Knowledge Exchange (e.g., IDVSCI [63]).
  - LLMs propose hypotheses based on GRAPHYP's adversarial routes.
  - Communities respond via search behavior metrics.

　　　　o　LLMs refine outputs using dual-diversity review (expert + adversarial evaluation) [63].

*Appendix A.3 Scientific Use Cases*

　　1. Climate Change Disputes

- Scenario: GRAPHYP identifies two cognitive communities:

    Community A: High Mass/Intensity; focused on anthropogenic models;
    Community B: High Variety; focused on natural variability.

- LLM Role:

    Generates comparative reports citing sources from each subnetwork;
    Highlights disputed metrics (e.g., temperature projections);
    Triggers assessor shifts and proposes alternative exploration paths [17].

    2. Genomics and Gene-Editing Controversies

- Scenario: Debates over CRISPR ethics surface via distinct search routes.
- LLM Role:

    Simulates peer review dialogue;
    Directs users to contrasting literature via GRAPHYP's bipartite hypergraphs.

*Appendix A.4. Social Network Applications*

　　1. Political Polarization

- GRAPHYP:

    Detects polarized cliques (e.g., vaccine-skeptic vs. pro-vaccine);
    Tracks shifts via content diversity.

- LLM:

    Offers personalized, bridging content across adversarial subnetworks.

    2. Misinformation Detection

- GRAPHYP:

    Flags disputed claims through query anomaly detection (e.g., "5G health risks").

- LLM:

    Generates counter-narratives through the following:

    Collaborative filtering: Links users to trusted sources;
    Personalized PageRank: Elevates high-centrality experts.

*Appendix A.5. Implementation Requirements*

　　Data Flow: Search logs → GRAPHYP subnetworks → LLM prompt engineering.
　　Evaluation: Dual-diversity review ensures balance between mainstream and **adversarial** perspectives [63].
　　This integration supports structured, real-time dialogues in scientific and social domains, advancing collective reasoning.

## Appendix B. Assessor Shifts for Scientific Dispute Resolution

　　GRAPHYP perspective to resolve scientific disputes with LLMs: Cognitive communities—groups of experts or stakeholders—can leverage assessor shifts (changes in evaluative stance during debates) to resolve disputes with LLM support.

*Appendix B.1. Capturing Assessor Shifts*

- Definition: Adjustments in how evidence or arguments are weighted in response to a new input.
- Modeling: GRAPHYP tracks shifts via changes in search behavior, source citation, and argumentative structure.

*Appendix B.2. Integrating LLMs for Arbitration*

- Conflict-Aware Reasoning**:** LLMs use frameworks like the Cognitive Alignment Framework to synthesize competing views through dual-process reasoning (heuristic + analytical).

    Example: In peer review, the LLM extracts arguments, maps conflicts, and generates a consensus meta-review.

- Bias Mitigation**:** LLMs can be trained to detect and counteract human biases (e.g., anchoring, conformity) for fairer outcomes.
- Multi-Agent Collaboration: Frameworks like RECONCILE simulate dialogical reasoning among LLM "agents," each representing distinct assessor positions. Through iterative voting, a reasoned consensus is formed.

*Appendix B.3. Key Benefits*

- Transparency**:** Documents how community positions evolve.
- Efficiency**:** Accelerates dispute resolution.
- Bias Reduction**:** Counters human and model biases.
- Scalability**:** Manages large-scale, multi-perspective debates beyond traditional peer review.

    By combining assessor shifts with LLM frameworks, scientific disputes can be approached more systematically and equitably, fostering transparent and robust knowledge production.